\documentclass[letterpaper, 10 pt, conference]{ieeeconf}  

\IEEEoverridecommandlockouts                              

\usepackage{graphics} 
\usepackage{amsmath} 
\usepackage{amssymb}  
\usepackage{adjustbox}
\usepackage{array}
\usepackage{cite}
\usepackage{multirow}
\usepackage{makecell}
\usepackage{hyperref}

\title{\LARGE \bf
Robust 6D Object Pose Estimation by Learning RGB-D Features
}

\author{Meng Tian$^{1}$, Liang Pan$^{1}$, Marcelo H Ang Jr$^{1}$ and Gim Hee Lee$^{2}$
\thanks{$^{1}$Department of Mechanical Engineering, National University of Singapore
        {\tt\small tianmeng@u.nus.edu, pan.liang@u.nus.edu, mpeangh@nus.edu.sg}}%
\thanks{$^{2}$Computer Vision and Robotic Perception Lab, Department of Computer Science, School of Computing, National University of Singapore
        {\tt\small gimhee.lee@nus.edu.sg}}%
}

\begin{document}

\maketitle
\thispagestyle{empty}
\pagestyle{empty}

\begin{abstract}

Accurate 6D object pose estimation is fundamental to robotic manipulation and grasping.
Previous methods follow a local optimization approach which minimizes the distance between closest point pairs to handle the rotation ambiguity of symmetric objects.
In this work, we propose a novel discrete-continuous formulation for rotation regression to resolve this local-optimum problem.
We uniformly sample rotation anchors in SO(3), and predict a constrained deviation from each anchor to the target, as well as uncertainty scores for selecting the best prediction.
Additionally, the object location is detected by aggregating point-wise vectors pointing to the 3D center.
Experiments on two benchmarks: LINEMOD and YCB-Video, show that the proposed method outperforms state-of-the-art approaches.
Our code is available at \url{https://github.com/mentian/object-posenet}.

\end{abstract}

\section{INTRODUCTION}

6D object pose estimation is an important foundation for various robotic tasks,
such as robotic manipulation and grasping \cite{correll2016analysis, tremblay2018deep}.
This problem is very challenging due to the varying illumination conditions, background clutters, and heavy occlusions between objects.
Objects that are symmetrical or textureless further aggravate this problem.
Additionally, a viable solution should be able to infer at real-time speed.

\begin{figure}[t]
    \centering
    \includegraphics[width=\columnwidth]{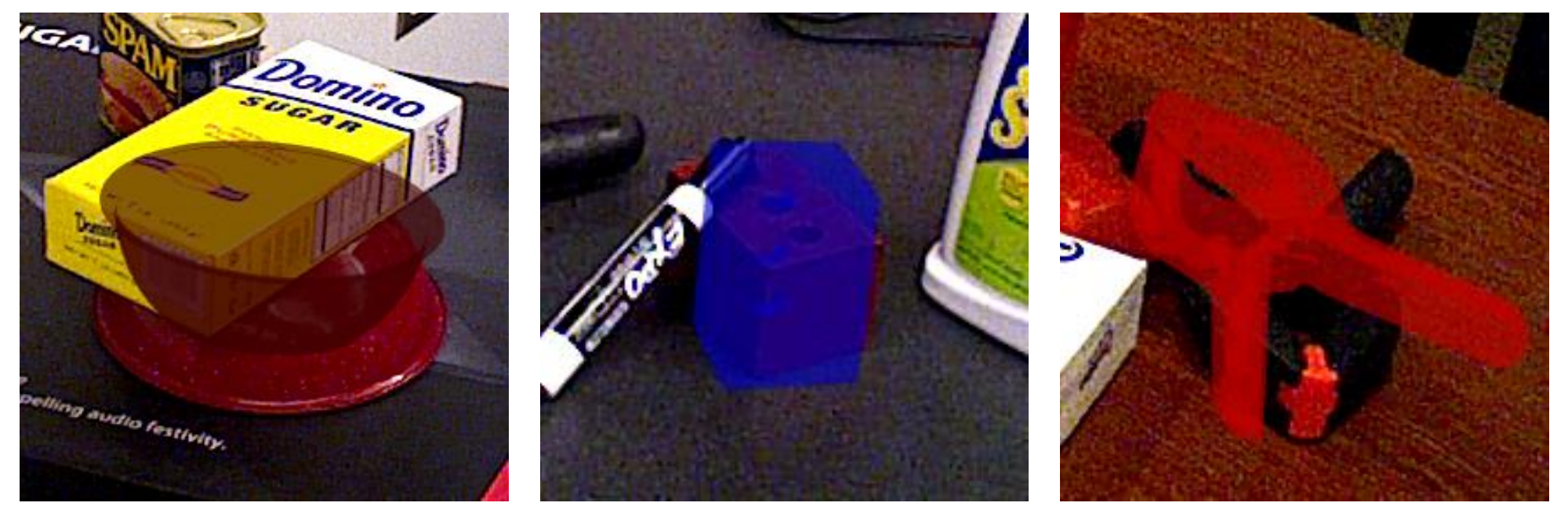}
    \caption{Some incorrect predictions from PoseCNN \cite{xiang2017posecnn} (left), DenseFusion \cite{wang2019densefusion} (middle), and our implementation of ShapeMatch-Loss (right).
    These errors are typical and commonly seen in the results of all three methods.
    It is strongly indicated that ShapeMatch-Loss suffers the local-optimum problem.}
    \label{fig:teaser}
\end{figure}

The recent explosion in 6D object pose estimation is arguably a result of the application of deep neural networks.
Many proposed deep networks \cite{kehl2017ssd, tekin2018real, xiang2017posecnn, sundermeyer2018implicit, peng2019pvnet, zakharov2019dpod, oberweger2018making, hu2018segmentation} only leverage RGB data, which are inherently sensitive to changing lighting conditions \cite{hodan2018bop} and object appearance variations \cite{sundermeyer2018implicit}.
To mitigate these problems, researchers start to take advantage of 3D geometric features and use RGB-D images for object pose estimation.
\cite{li2018unified} proposes a unified scalable framework with CNN architectures to exploit features from both RGB and depth images.
\cite{wang2019densefusion} introduces a PointNet-based \cite{qi2017pointnet} dense-fusion network that extracts geometric features from a point cloud and fuses with color features.
However, both of them fall short of fully resolving pose ambiguities caused by symmetric objects.
The former does not differentiate between symmetric and asymmetric objects and ignores the ambiguities.
The latter adopts a naive version of ShapeMatch-Loss \cite{xiang2017posecnn} which suffers the local-optimum problem (Fig. \ref{fig:teaser}) caused by minimizing the distance between closest point pairs.

A symmetric object can give rise to ambiguous rotation estimations, while it does not influence translation.
We argue that rotation and translation should be solved separately as opposed to jointly optimizing them.
To this end, we propose a novel end-to-end deep network which densely extracts features from RGB-D images and estimate rotation and translation in two separate branches.
For rotation, we initialize the estimation with uniformly sampled rotations (referred to as rotation anchors) in SO(3).
Given a target rotation, our network predicts a deviation from each rotation anchor to the target.
All the rotations anchor together cover the whole SO(3) space and each anchor is only responsible for a local region.
By local prediction and local optimization, our method resolves the local-optimum problem.
We also predict an uncertainty score for each anchor, which is used to select the best prediction during inference.
Inspired by \cite{novotny2018self, li2019usip}, the score is learnt in an self-supervised way by maximizing the conditional probability of the rotation error given the uncertainty score.
For translation, our network regresses a unit vector pointing towards the 3D object center at each point.
We integrate a RANSAC-based voting layer to aggregate all the vectors and select the best hypothesis as the final estimation.
Robustness of our method is twofold: first, by densely extracting RGB-D features our method is robust to changing lighting conditions and object appearance variations; and secondly, our regression strategies for both rotation and translation are robust.
To be specific, the proposed rotation regression consists of local region classification and constrained residual regression, which is more robust than a single regression strategy.
The RANSAC-based voting enforces local prediction such that it is robust to occlusions and clutters.

We evaluate our method on two widely-used benchmarks for 6D pose estimation: LINEMOD dataset \cite{hinterstoisser2012model} and YCB-Video \cite{xiang2017posecnn} dataset.
Our method achieves superior performance on both datasets over the state-of-the-art methods.
We also demonstrate its robustness to heavy occlusions, background clutters and varying lighting conditions.

In summary, our key contributions are as follows:
\begin{itemize}
	\item We propose a novel deep network for 6D pose estimation, which robustly predicts rotation and translation from densely extracted RGB-D features.
	\item We propose a discrete-continuous formulation to resolve the local-optimum problem of ShapeMatch-Loss and handle rotation ambiguities.
	\item Our network outperforms the state-of-the-art results on both LINEMOD (ADD: 92.8\% vs. 86.3\%) and YCB-Video (ADD: 83.8\% vs. 79.2\%).
\end{itemize}

\section{RELATED WORK}

\textbf{RGB methods.}
Traditionally, object pose estimation from single RGB image is tackled by matching local features \cite{lowe2004distinctive, lowe1999object}.
However, sufficient textures are required to compute the features.
To handle textureless objects, \cite{brachmann2014learning} and \cite{brachmann2016uncertainty} propose to directly regress 3D object coordinates.
But the sampling of pose hypothesis and refinement are very time-consuming.
Recent methods apply deep learning techniques to this task.
\cite{kehl2017ssd} extends an 2D object detection network \cite{liu2016ssd} to predict object's identity, 2D bounding box and a discretized orientation.
\cite{tekin2018real} and \cite{rad2017bb8} first detect the keypoints of the object and then solve a Perspective-n-Point (PnP) problem for pose estimation.
These methods predict from the global feature of the object.
Hence, performance on occluded objects are not satisfactory.
To address the occlusion problem, dense prediction methods \cite{xiang2017posecnn, oberweger2018making, hu2018segmentation, peng2019pvnet, zakharov2019dpod} are proposed.
\cite{oberweger2018making} and \cite{hu2018segmentation} estimate a pose at each visible local patch independently, and accumulate results over all patches.
\cite{xiang2017posecnn} and \cite{peng2019pvnet} predict a unit vector at each pixel pointing towards keypoints.
\cite{zakharov2019dpod} estimates a dense 2D-3D correspondence map between input image and object models.
Among them, \cite{peng2019pvnet} and \cite{zakharov2019dpod} achieve top performances.
But they are still inferior to RGB-D based methods in terms of detection accuracy.

\textbf{RGB-D methods.}
LINEMOD \cite{hinterstoisser2012model} is considered as a seminal work, after which all methods are benchmarked.
They estimate object pose by matching templates consisting of color gradient and surface normal features.
Their templates are holistic, thus cannot handle occluded objects.
\cite{brachmann2014learning, krull2015learning, michel2017global} regress a 3D object coordinate for each pixel on image, from which pose hypotheses are generated and further refined.
These methods rely on hand-crafted features which limits their performances.
Both \cite{tejani2014latent} and \cite{kehl2016deep} use locally-sampled RGB-D patches to cast votes for 6D pose.
\cite{tejani2014latent} proposes a Latent-Class Hough Forest which stores center position and rotation at its leaf nodes.
\cite{kehl2016deep} employs a convolutional auto-encoder to extract descriptors, and match to the nearest neighbor in a codebook.
\cite{li2018unified} and \cite{wang2019densefusion} are two recent works that predict 6D object pose from RGB-D images in an end-to-end way.
\cite{li2018unified} processes color image and depth image separately, then concatenates two feature maps along feature dimension.
The global fusion does not make full use of geometric information.
In contrast, we follow a dense-fusion way as in \cite{wang2019densefusion} and \cite{dai20183dmv}.
\cite{wang2019densefusion} is the most similar work to us.
However, we propose a novel discrete-continuous formulation for rotation prediction to resolve the local-optimum issue.
We also extend the 2D RANSAC-based voting to 3D space for estimating translation, which is a much more robust method than single point prediction.

\section{METHOD}

The task of 6D object pose estimation is to detect each object instance and meanwhile estimate a rigid transformation from the object coordinate frame to the camera coordinate frame.
Specifically, this rigid transformation is represented by a rotation $\mathbf{R} \in SO(3)$ and a translation $\mathbf{t} \in R^3$.

\subsection{Overview}

\begin{figure*}[t]
    \centering
    \includegraphics[width=\textwidth]{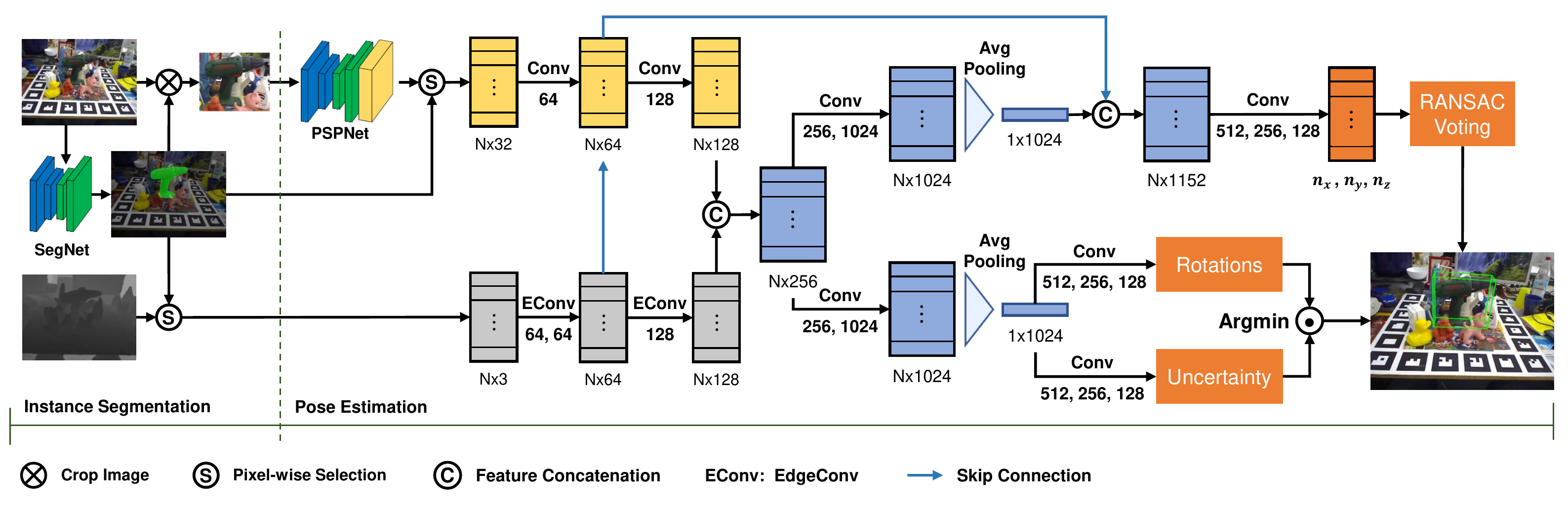}
    \caption{\textbf{Overview of our pipeline.} Instance mask is first obtained through a semantic segmentation network (SegNet). With the obtained mask, our pose estimation network  extracts a point cloud from depth image and crops a tight image window from color image as inputs. Color features and geometric features are densely concatenated, from which rotation and translation of the object are predicted in two separate branches.}
    \label{fig:overview}
\end{figure*}

Fig. \ref{fig:overview} overviews the proposed two-stage pipeline: instance segmentation and pose estimation.
The first stage is to detect each object instance on the color image and obtain a foreground mask (e.g. instance segmentation).
Considering that each object only appears at most once on the images we evaluate on, we adopt the well-developed SegNet \cite{badrinarayanan2015segnet}, which is originally designed for semantic segmentation, as our pre-processing network.
If there are multiple instances of same object category presented on an image, this part could be straightforwardly replaced by an instance segmentation network, such as \cite{He_2017_ICCV} and \cite{liu2018path}.
With the obtained mask, we crop an image window which tightly encloses the object, and extract a point cloud from depth image as inputs to the second-stage pose estimation network.
Our network concatenates densely extracted color and geometric features in a pixel-wise manner, from which rotation and translation are predicted in two separate branches.
For rotation branch, we propose a robust discrete-continuous regression scheme to handle ambiguities caused by symmetrical objects.
The translation branch regresses a unit vector pointing towards the object center in 3D space at each point, and aggregates these vectors through RANSAC-based voting.

\subsection{Rotation}

Symmetric objects will cause convergence problem during training, since there are multiple rotation labels corresponding to an identical appearance.
To alleviate this problem, ShapeMatch-Loss \cite{xiang2017posecnn} is proposed and defined as
\begin{equation}
 L = \begin{cases}
  \displaystyle\frac{1}{M} \displaystyle\sum_{\mathbf{x}_1 \in \mathcal{M}} \displaystyle\min_{\mathbf{x}_2 \in \mathcal{M}} \| \mathbf{R} \mathbf{x}_1 - \mathbf{\Tilde{R}} \mathbf{x}_2) \|_2 & \text{if symmetric} \,,\\
  \displaystyle\frac{1}{M} \displaystyle\sum_{\mathbf{x} \in \mathcal{M}} \| \mathbf{R} \mathbf{x} - \mathbf{\Tilde{R}} \mathbf{x}) \|_2 & \text{if asymmetric} \,;
 \end{cases}
\end{equation}
where $M$ is the number of points in object model $\mathcal{M}$, $\mathbf{\tilde{R}}$ and $\mathbf{R}$ are the ground truth and prediction.
Ambiguous labels of symmetric objects become equivalent under this loss function.
However, just like ICP, ShapeMatch-Loss is only a local optimization for symmetric objects.
It will slow down the training process and converge to sub-optimal solution in the worst-case scenario.

We resolve this local-optimum issue by initializing our predictions with rotation anchors covering the whole $SO(3)$ space.
Specifically, instead of directly estimating $\mathbf{R}$, our network predicts a deviation $\Delta \mathbf{R}_i$ at rotation anchor $\mathbf{\hat{R}}_i$:
\begin{equation}
\mathbf{R}_i = \Delta \mathbf{R}_i \mathbf{\hat{R}}_i\, ,
\end{equation}
where $\mathbf{R}_i$ is the total prediction at $i$-th rotation anchor.
$\Delta \mathbf{R}_i$ is constrained to be within the vicinity of $\mathbf{\hat{R}}_i$, such that each anchor responds for a local region.
We investigate three discrete subgroups of $SO(3)$ as rotation anchors, namely tetrahedral group ($N$=12), octahedral group ($N$=24), and icosahedral group ($N$=60) \cite{yan2012almost}, with $N$ being the number of rotations in the corresponding group.
Each of these three subgroups is a uniform sampling of $SO(3)$.
In addition to $\Delta \mathbf{R}_i$, our network also predicts an uncertainty score $\sigma_i \in [0, 1]$ for each anchor.
$\sigma_i$ is proportional to the distance between $\mathbf{R}_i$ and ground-truth rotation after the training is converged.
During inference, prediction of the anchor with smallest uncertainty score is selected as the final output.

For each prediction $\mathbf{R}_i$, we define the conditional probability of the normalized ShapeMatch-Loss $d_i=L_i/d$ given $\sigma_i$ as
\begin{equation}
p(d_i|\sigma_i) = \frac{1}{\sigma_i}\exp(-\frac{d_i}{\sigma_i})\,,
\end{equation}
where $d$ is the object diameter, $L_i$ is the ShapeMatch-Loss of $\mathbf{R}_i$.
For a certain $d_i$, maximum of $p$ is achieved when $\sigma_i = d_i$. Maximizing $p$ equals to minimizing $d_i$ and setting $\sigma_i$ to $d_i$.
Therefore, optimal parameters of the network can be obtained by minimizing the probabilistic loss
\begin{equation}
\label{eqn:loss_r}
  \begin{split}
    L_R & = -\sum_i \ln{p(d_i|\sigma_i)} \\
    & = \sum_i \ln{\sigma_i} + \frac{d_i}{\sigma_i} \,.
  \end{split}
\end{equation}
The regularization loss which constrains the range of $\Delta \mathbf{R}_i$ is
\begin{equation}
  L_{reg} = \sum_i \max(0, \, \max_{j \neq i} \langle  \mathbf{q}_i, \mathbf{\hat{q}}_j \rangle - \langle \mathbf{q}_i, \mathbf{\hat{q}}_i \rangle) \, ,
\end{equation}
where $\mathbf{q}_i$ and $\mathbf{\hat{q}}_i$ are the quaternion representations of $\mathbf{R}_i$ and $\mathbf{\hat{R}}_i$.

The proposed regression strategy has another benefit of improving the regressing accuracy.
It is actually a practice of reformulating a continuous regression problem as a joint classification and residual regression problem.
This kind of reformulation has been widely proven to be effective \cite{ren2015faster, mousavian20173d} and indeed verified in our experiments.

\textbf{Remarks:}
\cite{li2018unified} also uses a discrete-continuous formulation for regressing rotation.
However, they do not enforce local prediction for each anchor.
Their classification over the anchors is mutually exclusive, such that ambiguous labels of symmetric objects will confuse the classifier.
Our formulation can be viewed as independently conducting binary classification for each anchor, which is more suitable for symmetric objects,
Loss function of \cite{wang2019densefusion} could be derived from our probabilistic loss (Eq.\ref{eqn:loss_r}) by a reparameterization trick.
The confidence value $c$ in their loss function is the inverse of our uncertainty score.
Hence, optimizing $c$ leads to unbounded prediction.
Their loss equally penalizes the distance error of objects of different sizes, thus has a bias towards large objects.
While our loss normalizes the error with the corresponding object size.
\cite{manhardt2018explaining} proposes to resolve the rotation ambiguity by generating multiple hypotheses (e.g. 5).
Compared with their solution, our formulation dynamically selects the number of hypotheses (up to the total number of anchors) according to the uncertainty scores.

\subsection{Translation}

In order to make use of the complementary depth information, we extend a RANSAC-based voting method \cite{xiang2017posecnn, peng2019pvnet} from 2D to 3D space.
Point-wise prediction focuses more on local features, which makes our prediction robust to foreground occlusions, background clutters and segmentation noise in the first stage.
Suppose that the object center is $\mathbf{c} \in R^3$.
At each selected point $\mathbf{p}$, our network predicts a unit vector $\mathbf{v}$ representing the direction from that point to the center:
\begin{equation}
\mathbf{v} = \frac{\mathbf{c} - \mathbf{p}}{\| \mathbf{c} - \mathbf{p} \|_2}\,.
\end{equation}
Randomly sampled two points and their associated vectors define two 3D lines.
However, lines do not necessarily intersect in 3D space.
To mitigate this problem, a hypothesis $\mathbf{h}$ is generated by taking the mid-point of the shortest line segment between these two lines.
Votes are accumulated over all points.
A point is considered as an inlier if $(\mathbf{h} - \mathbf{p})^T \mathbf{v}/ \| \mathbf{h} - \mathbf{p} \|_2 \geq \theta$, where $\theta$ is a threshold and set to 0.99 for all experiments.
Final estimate is set to the point closest to all the lines defined by the inliers of the hypothesis with highest voting score.
During training, we employ smooth L1 loss ($L_t$) for learning the vectors.

\subsection{Network Architecture}

Our network has two separate branches for color feature and geometric feature extraction.
Features from two different modalities are densely concatenated, from which translation and rotation are predicted independently.
The translation branch enforces dense prediction.
Therefore, we concatenate high-level global feature with low-level color and geometric features (shown as skip connections in Fig. \ref{fig:overview}).
While the rotation branch only leverages the global average-pooled feature.

We use a fully convolutional network modified from PSPNet \cite{zhao2017pyramid} to densely extract color features.
With ResNet-18 \cite{he2016deep} as backend, up-sampling layers are applied after pyramid pooling to restore the size of feature map.
For arbitrary-sized image crop, network will output a feature map of same size.
Then, those features belong to foreground pixels are selected, and lifted to the same dimension with geometric features by two 1x1 convolution layers.
Local geometric structure has been proven useful for reasoning on point cloud \cite{wang2018dynamic}.
We recurrently apply edge convolution operation (EdgeConv) proposed in \cite{wang2018dynamic} to extract geometric features.
The obtained dense color and geometric features are concatenated based on pixel-to-point correspondences.
Observing that color feature and geometric feature play different roles in estimating translation and rotation, it is better to separate two prediction branches and let the network learn to balance two-modal information, rather than constraining them to share one global feature.
Our experiment validates this particular design of network architecture.

\section{EXPERIMENTS}

\subsection{Datasets}

\textbf{LINEMOD} \cite{hinterstoisser2012model} is the standard and most popular benchmark for 6D object pose estimation.
This dataset has 13 texture-less objects placed in heavily cluttered scenes, of which two are symmetric.
Following \cite{brachmann2016uncertainty, rad2017bb8} and \cite{wang2019densefusion}, training images are selected such that relative orientation between two images are larger than a threshold (e.g. $15^{\circ}$), and viewpoints of the upper hemisphere are regularly covered.
About $15\%$ of all images are selected for training.
The rest are used for testing.

\textbf{YCB-Video} \cite{xiang2017posecnn} consists of 92 real video sequences collected for 21 objects from YCB objects \cite{calli2015ycb} and 80k synthetically rendered images.
2949 key frames are extracted from 12 video sequences for testing, while the remaining sequences and synthetic images can be used for training.
This dataset is extremely challenging due to varying light conditions, presence of heavy occlusions, and significant image noise.

\subsection{Evaluation Metrics}

We report the most widely used pose error in 3D space, which is referred to as ADD metric \cite{hinterstoisser2012model}.
ADD is defined as the average distance between model vertices transformed with estimated pose $(\mathbf{R}, \mathbf{t})$ and ground-truth pose $(\mathbf{\tilde{R}}, \mathbf{\tilde{t}})$:
\begin{equation}
\text{ADD} = \frac{1}{M} \sum_{\mathbf{x} \in \mathcal{M}} \| (\mathbf{R} \mathbf{x} + \mathbf{t}) - (\mathbf{\tilde{R}} \mathbf{x} + \mathbf{\tilde{t}}) \|_2 \,,
\end{equation}
where $\mathbf{x} \in \mathcal{M}$ denotes the 3D model point and $M$ is the total number of points.
A predicted 6D pose is considered to be correct if ADD is less than a threshold (e.g. $10\%$ of object diameter).
For symmetric objects, the error is calculated as the average distance to the closest model point \cite{hodavn2016evaluation}:
\begin{equation}
\text{ADD-S} = \frac{1}{M} \sum_{\mathbf{x_1} \in \mathcal{M}} \min_{\mathbf{x_2} \in \mathcal{M}} \| (\mathbf{R} \mathbf{x_1} + \mathbf{t}) - (\mathbf{\tilde{R}} \mathbf{x_2} + \mathbf{\tilde{t}}) \|_2 \,.
\end{equation}
ADD(-S) combines two metrics and means we use ADD metric for asymmetric objects and ADD-S metric for symmetric objects during evaluation.
AUC (area under accracy-threshold curve) is a commonly used metric on YCB-Video dataset \cite{xiang2017posecnn, wang2019densefusion, li2018unified, oberweger2018making}.
We also report AUC in terms of ADD-S metric within a range of $[0, 0.1]$.

\subsection{Implementation Details}

The total loss function is $L = L_R + \lambda_1 L_{reg} + \lambda_2 L_t$.
We empirically find that $\lambda_1 = 2$ and $\lambda_2 = 5$ are good choices.
Initial learning rate is set to 0.0001, and decayed by a rate of 0.6 and 0.5 when ADD(-S) metric on a validation set is less than 0.016 and 0.013 respectively.
We randomly sample 500 and 1,000 foreground points for each object in LINEMOD and YCB-Video datasets as inputs.
3D RANSAC-voting layer is implemented using CUDA.
Each iteration parallelly generates 128 hypotheses.
The maximum number of iterations is 20, resulting in 2,560 hypotheses altogether at most.
Early stop requires the probability of success above 0.99.
Probability of choosing an inlier is set to the maximum inlier ratio of currently generated hypotheses.

\begin{table}[t]
\caption{Ablation Studies: accuracies on LINEMOD dataset in terms of ADD(-S) metric are reported.}
\label{table:ablation}
\centering
\setlength{\tabcolsep}{10pt} 
\renewcommand{\arraystretch}{1.2} 
\begin{adjustbox}{max width=\columnwidth}
\begin{tabular}{| r |  c |}
 \hline
 Architecture & Accuracy \\
 \hline
 DenseFusion \cite{wang2019densefusion} & 86.20 \\
 RANSAC voting          & 89.42 \\
 local geoemtry         & 90.36 \\
 60 anchors             & \textbf{92.77} \\
 shared global feature  & 91.89 \\
 w/o regularization     & 84.71 \\
 24 anchors             & 89.89 \\
 12 anchors             & 87.86 \\
 \hline
\end{tabular}
\end{adjustbox}
\end{table}

\begin{table*}[t]
\caption{Comparison on LINEMOD dataset. Objects marked with * are considered to be symmetric.}
\label{table:linemod_comparison}
\centering
\renewcommand{\arraystretch}{1.2} 
\begin{adjustbox}{max width=0.9\textwidth}
\begin{tabular}{ | r | c  c  c | c  c | c  c | c |}
 \hline
 & \multicolumn{3}{c|}{RGB w/o refinement} & \multicolumn{2}{c|}{RGB w/ refinement} & \multicolumn{2}{c|}{RGB-D w/o refinement} & {RGB-D w/ refinement} \\
 \hline
 Method & PoseCNN \cite{xiang2017posecnn} & PVNet \cite{peng2019pvnet} & DPOD \cite{zakharov2019dpod} & DeepIM \cite{Li_2018_ECCV} & DPOD+ \cite{zakharov2019dpod} & Per-Pixel DF \cite{wang2019densefusion} & \textbf{Ours} & Iterative DF \cite{wang2019densefusion} \\
 \hline
 ape            & 21.62 & 43.62 & 53.28 & 77.0 & 87.73 & 79.5 & 85.03 & 92.3 \\
 benchvise      & 81.80 & 99.90 & 95.34 & 97.5 & 98.45 & 84.2 & 95.54 & 93.2 \\
 cam            & 36.57 & 86.86 & 90.36 & 93.5 & 96.07 & 76.5 & 91.27 & 94.4 \\
 can            & 68.80 & 95.47 & 94.10 & 96.5 & 99.71 & 86.6 & 95.18 & 93.1 \\
 cat            & 41.82 & 79.34 & 60.38 & 82.1 & 94.71 & 88.8 & 93.61 & 96.5 \\
 driller        & 63.51 & 96.43 & 97.72 & 95.0 & 98.80 & 77.7 & 82.56 & 87.0 \\
 duck           & 27.23 & 52.58 & 66.01 & 77.7 & 86.29 & 76.3 & 88.08 & 92.3 \\
 eggbox*        & 69.58 & 99.15 & 99.72 & 97.1 & 99.91 & 99.9 & 99.90 & 99.8 \\
 glue*          & 80.02 & 95.66 & 93.83 & 99.4 & 96.82 & 99.4 & 99.61 & 100.0 \\
 holepuncher    & 42.63 & 81.92 & 65.83 & 52.8 & 86.87 & 79.0 & 92.58 & 92.1 \\
 iron           & 74.97 & 98.88 & 99.80 & 98.3 & 100.0 & 92.1 & 95.91 & 97.0 \\
 lamp           & 71.11 & 99.33 & 88.11 & 97.5 & 96.84 & 92.3 & 94.43 & 95.3 \\
 phone          & 47.74 & 92.41 & 74.24 & 87.7 & 94.69 & 88.0 & 93.56 & 92.8 \\
 \hline
 Average & 55.95 & \textbf{86.27} & 82.98 & 88.6 & \textbf{95.15} & 86.2 & \textbf{92.87} & \textbf{94.3} \\
 \hline
\end{tabular}
\end{adjustbox}
\end{table*}

\subsection{Ablation Analysis}

Extensive ablation studies are conducted to compare different design choices.
Table \ref{table:ablation} summaries the evaluation results.
We incrementally modify the network of DenseFusion \cite{wang2019densefusion} and add in our innovations.
First of all, we explore the 3D RANSAC-based voting technique for translation estimation.
DenseFusion predicts an offset to object center at each point.
While our network outputs an unit vector pointing to the center, which is equivalent to the normalized offset.
The overall detection accuracy increases by 3.2\%, showing that 3D voting yields more accurate results due to the robustness of RANSAC.
Then, we validate the benefit of incorporating local geometry by applying EdgeConv \cite{wang2018dynamic} layer.
The accuracy increases by another 0.9\%.
To analyze the proposed discrete-continuous formulation for rotation, we replace the per-point prediction in "local geometry" variant with our anchor-based prediction.
Correspondingly, ShapeMatch-Loss is also replaced by our probabilistic loss during training.
When there are 60 anchors, our best performance on LINEMOD dataset is achieved.
Compared with the "local geometry", our accuracy is 2.4\% higher.

We also train the network with 24 and 12 anchors respectively.
The accuracy decreases considerably along with the number of anchors.
Additionally, we test the case in which rotation and translation branches share one global feature.
This single modification makes the accuracy drops by 0.9\% from our best model, proving that rotation and translation pay different attentions to color and geometric features.
If we remove the regularization on the range of deviations, which means at each anchor our network can predict arbitrary rotation in $SO(3)$, the accuracy decreases significantly (e.g. 8.1\%).
Local prediction is empirically proven to be the key ingredient of our discrete-continuous formulation.
Last but not the least, we re-implement the regression method proposed in \cite{li2018unified} and replace the corresponding part in our best model.
On YCB-Video dataset, the accuracy decreases from 83.8\% to 71.1\%.
The reasons are twofold: their mutually exclusive classification and loss function do not differentiate between symmetric and asymmetric objects; they do not enforce local prediction for each anchor.

\begin{table}[t]
\caption{AUC in terms of ADD-S metric on YCB-Video dataset. * indicates symmetric object.}
\label{table:ycb_auc}
\centering
\renewcommand{\arraystretch}{1.2} 
\setlength{\tabcolsep}{2pt}
\begin{adjustbox}{max width=\columnwidth}
\begin{tabular}{ | r | c  c  c | c  c  c |}
 \hline
 {} & \multicolumn{3}{c|}{RGB-D} & \multicolumn{3}{c|}{RGB-D w/ Refinement} \\
 \hline
 Method & \makecell{MCN \\ \cite{li2018unified}} & \makecell{Per-Pixel \\ DF \cite{wang2019densefusion}} & \textbf{Ours} & \makecell{PoseCNN \\ +ICP \cite{xiang2017posecnn}} & \makecell{MCN+ \\ ICP \cite{li2018unified}} & \makecell{Iterative \\ DF \cite{wang2019densefusion}} \\
 \hline
 002\_master\_chef\_can     & 89.4  & 95.2  & 93.9  & 95.8  & 96.0  & 96.4 \\
 003\_cracker\_box          & 85.4  & 92.5  & 92.9  & 91.8  & 88.7  & 95.5 \\
 004\_sugar\_box            & 92.7  & 95.1  & 95.4  & 98.2  & 97.3  & 97.5 \\
 005\_tomato\_soup\_can     & 93.2  & 93.7  & 93.3  & 94.5  & 96.5  & 94.6 \\
 006\_mustard\_bottle       & 96.7  & 95.9  & 95.4  & 98.4  & 97.7  & 97.2 \\
 007\_tuna\_fish\_can       & 95.1  & 94.9  & 94.9  & 97.1  & 97.6  & 96.6 \\
 008\_pudding\_box          & 91.6  & 94.7  & 94.0  & 97.9  & 86.2  & 96.5 \\
 009\_gelatin\_box          & 94.6  & 95.8  & 97.6  & 98.8  & 97.6  & 98.1 \\
 010\_potted\_meat\_can     & 91.7  & 90.1  & 90.6  & 92.8  & 90.8  & 91.3 \\
 011\_banana                & 93.8  & 91.5  & 91.7  & 96.9  & 97.5  & 96.6 \\
 019\_ptcher\_base          & 93.8  & 94.6  & 93.1  & 97.8  & 96.6  & 97.1 \\
 021\_bleach\_cleanser      & 92.9  & 94.3  & 93.4  & 96.8  & 96.4  & 95.8 \\
 024\_bowl*                 & 82.6  & 86.6  & 92.9  & 78.3  & 76.0  & 88.2 \\
 025\_mug                   & 95.3  & 95.5  & 96.1  & 95.1  & 97.3  & 97.1 \\
 035\_power\_drill          & 88.2  & 92.4  & 93.3  & 98.0  & 95.9  & 96.0 \\
 036\_wood\_block*          & 81.5  & 85.5  & 87.6  & 90.5  & 93.5  & 89.7 \\
 037\_scissors              & 87.3  & 96.4  & 95.7  & 92.2  & 79.2  & 95.2 \\
 040\_large\_marker         & 90.2  & 94.7  & 95.6  & 97.2  & 98.0  & 97.5 \\
 051\_large\_clamp*         & 91.5  & 71.6  & 75.4  & 75.4  & 94.0  & 72.9 \\
 052\_extra\_large\_clamp*  & 88.0  & 69.0  & 73.0  & 65.3  & 90.7  & 69.8 \\
 061\_foam\_brick*          & 93.2  & 92.4  & 94.2  & 97.1  & 96.5  & 92.5 \\
 \hline
 Average & 90.6 & 91.2 & \textbf{91.8} & 93.0 & \textbf{93.3} & 93.1 \\
 \hline
\end{tabular}
\end{adjustbox}
\end{table}

\subsection{Comparison on LINEMOD}

We compare our method with the state-of-the-art object pose detectors \cite{xiang2017posecnn, peng2019pvnet, zakharov2019dpod, wang2019densefusion} on the LINEMOD dataset.
Results are reported in Table \ref{table:linemod_comparison}.
Of all those methods without pose refinement, PVNet \cite{peng2019pvnet} and Per-Pixel DF \cite{wang2019densefusion} are the state of the art on LINEMOD dataset.
PVNet utilizes a similar voting technique to detect keypoints on color image, and pose is obtained through solving a Perspective-n-Point problem.
Although Per-Pixel DF uses the depth information, their performance is only comparable to PVNet.
Comparing with them, our detection accuracy is 6.6\% higher, and close to the results after pose refinement.
Especially, our method can handle small objects very well, such as "ape" (5.5\% higher), "duck" (11.8\% higher) and "holepuncher" (13.6\% higher).
These small textureless objects are inherently difficult for those methods based on establishing dense or sparse 2D-3D correspondences, even after pose refinement.

\begin{figure}[t]
    \centering
    \includegraphics[width=\columnwidth]{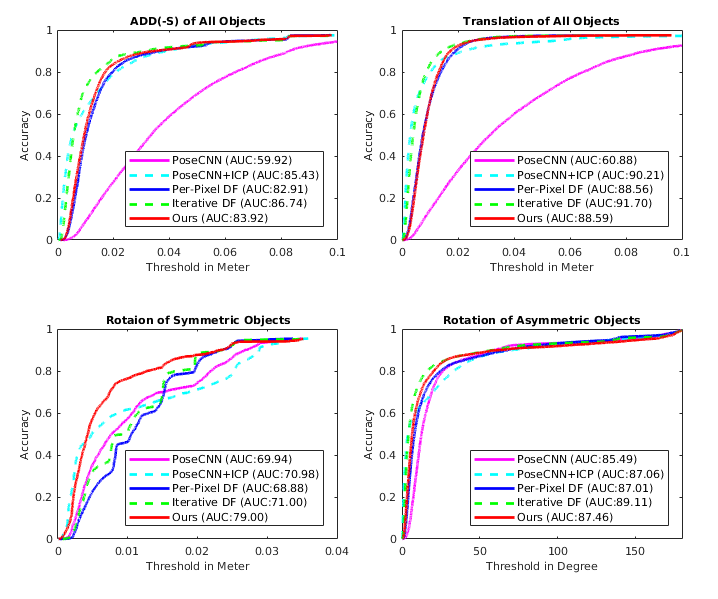}
    \caption{\textbf{Accuracy-threshold curves.} Top-left: pose error of all objects measured by ADD(-S) metric. Top-right: translation error of all objects. Bottom-left: rotation error of all symmetric objects in terms of ADD-S metric. Bottom-right: rotation error of all asymmetric objects in terms of relative rotation angle. Dashed line indicates that methods have pose refinement.}
    \label{fig:ycb_auc}
\end{figure}

\subsection{Comparison on YCB-Video}

Table \ref{table:ycb_auc} and \ref{table:ycb_add} compare our method with \cite{xiang2017posecnn, oberweger2018making, hu2018segmentation, wang2019densefusion, li2018unified} on the more challenging YCB-Video dataset.
For fair comparisons, we use the segmentation masks from \cite{xiang2017posecnn} and \cite{wang2019densefusion}.
In Table \ref{table:ycb_auc}, our average AUC in terms of ADD-S metric is 0.6\% higher than the state-of-the-art method (e.g. Per-Pixel DF \cite{wang2019densefusion}).
Since the ADD-S metric computes the average distances between closest point pairs, regardless of whether the object is symmetric or not, the ability of different detectors is not well reflected through this evaluation.
Therefore, we also compare the detection accuracy in terms of ADD(-S) metric.
our method achieves a notably improvement over Per-Pixel DF (4.6\% higher).
In general, accuracies of RGB-D based methods are much higher than RGB-only methods, demonstrating the robustness induced by geometric features.

\begin{figure}[t]
    \centering
    \includegraphics[width=\columnwidth]{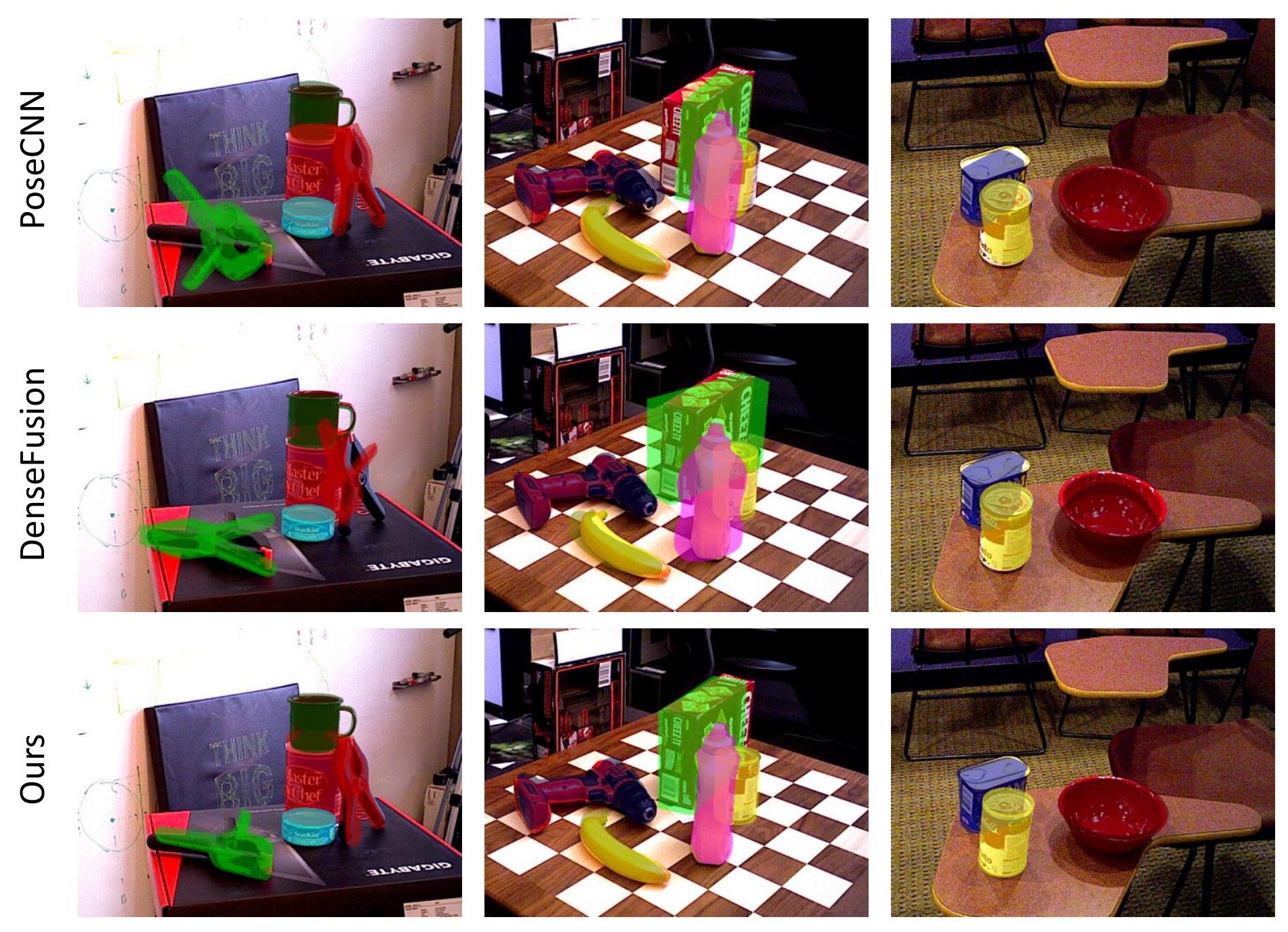}
    \caption{Qualitative results on YCB-Video dataset. Some cases in which our method (bottom row) can accurately predict the poses, while PoseCNN \cite{xiang2017posecnn} (top row) and Per-Pixel DF \cite{wang2019densefusion} (middle row) gives inaccurate or local optimal predictions.}
    \label{fig:ycb_qualitative}
\end{figure}

Fig.\ref{fig:ycb_auc} decouples the pose errors and provides detailed analysis on rotation and translation respectively.
Our method is marginally better on estimating translation (top-right) and rotation of asymmetric objects (bottom-right).
However, we achieve superior performance on estimating rotation of symmetric objects (bottom-left).
Both PoseCNN and DenseFusion naively implement the ShapeMatch-Loss.
Our remarkably higher AUC on symmetric objects shows that the proposed discrete-continuous formulation is able to handle the local-optimum problem of ShapeMatch-Loss.
Furthermore, pose refinement can effectively refine a rotation only when the initial prediction is near the ground truth.
For large errors which usually caused by local optimal predictions, the refinement performs no better than its baseline.
To this end, it is important to purposely deal with the local-optimum problem.
Fig.\ref{fig:ycb_qualitative} shows some qualitative comparisons between our method, PoseCNN and Per-Pixel DF.

\subsection{Runtime Analysis}
On single Nvidia GPU (GTX 1080Ti), our pose estimation network infers at 0.04s per instance.
Instance segmentation at the first stage takes about 0.03s per frame.
By concurrently running multiple pose estimation networks (one for each object instance), our method could run at about 14 FPS (0.07s per frame), which is very  promising for real-time applications.

\begin{table}[t]
\caption{Average accuracies of our method and baseline methods on YCB-Video dataset in terms of ADD(-S) metric.}
\label{table:ycb_add}
\centering
\setlength{\tabcolsep}{2pt} 
\renewcommand{\arraystretch}{1.6} 
\begin{adjustbox}{max width=\columnwidth}

\begin{tabular}{| r | c  c  c | c  c | c  c |}
 \hline
 {} & \multicolumn{3}{c|}{RGB} & \multicolumn{2}{c|}{RGB-D} & \multicolumn{2}{c|}{RGB-D w/ Refinement} \\
 \hline
 Method & \makecell{PoseCNN \\ \cite{xiang2017posecnn}} & \makecell{Heatmaps \\ \cite{oberweger2018making}} & \makecell{Seg-Driven \\ \cite{hu2018segmentation}} & \makecell{Per-Pixel \\ DF \cite{wang2019densefusion}} & \textbf{Ours} & \makecell{PoseCNN \\ +ICP \cite{xiang2017posecnn}} & \makecell{Iterative  \\ DF \cite{wang2019densefusion}} \\
 \hline
 Accuracy & 22.9 & \textbf{53.1} & 39.0 & 79.2 & \textbf{83.8} & 73.7 & \textbf{85.6} \\
 \hline

\end{tabular}
\end{adjustbox}
\end{table}

\section{CONCLUSION}
We present a deep learning method for 6D object pose estimation.
Specially, a novel discrete-continuous formulation is proposed for rotation to handle the local-optimum problem during training.
In the future, we would consider to use the uncertainty values for pose refinement, as well as introduce them to grasping algorithms.
Another extension is to test whether our approach can achieve similarly good results if trained purely on synthetic data.


\section*{ACKNOWLEDGMENT}

This work was supported in part by the Singapore MOE Tier 1 grant R-252-000-A65-114 and the National Research Foundation, Prime Ministers Office, Singapore, under its CREATE programme, Singapore-MIT Alliance for Research and Technology (SMART) Future Urban Mobility (FM) IRG.

\addtolength{\textheight}{-1.4cm}
\clearpage
\bibliographystyle{IEEEtran}
\bibliography{example}


\end{document}